\documentclass{article}

\usepackage[nonatbib, final]{nips_2017}

\usepackage[utf8]{inputenc} 
\usepackage[T1]{fontenc}    
\usepackage{hyperref}       
\usepackage{url}            
\usepackage{booktabs}       
\usepackage{amsfonts}       
\usepackage{nicefrac}       
\usepackage{microtype}      

\usepackage{graphicx}
\usepackage{caption} 
\usepackage{amsmath}
\usepackage{bm}

\usepackage[noadjust]{cite}

\title{Ligand Pose Optimization with Atomic Grid-Based Convolutional Neural Networks}

\author{
  Matthew Ragoza \\
  Computational \& Systems Biology \\
  University of Pittsburgh \\
  Pittsburgh, PA 15213 \\
  \texttt{mtr22@pitt.edu} \\
  \And
  Lillian Turner \\
  Department of Computer Science \\
  University of Pittsburgh \\
  Pittsburgh, PA 15213 \\
  \texttt{lmt72@pitt.edu} \\
  \And
  David Ryan Koes \\
  Computational \& Systems Biology \\
  University of Pittsburgh \\
  Pittsburgh, PA 15213 \\
  \texttt{dkoes@pitt.edu} \\
}

\begin{document}

\maketitle

\begin{abstract}
Docking is an important tool in computational drug discovery that aims to predict the binding pose of a ligand to a target protein through a combination of pose scoring and optimization. A scoring function that is differentiable with respect to atom positions can be used for both scoring and gradient-based optimization of poses for docking. Using a differentiable grid-based atomic representation as input, we demonstrate that a scoring function learned by training a convolutional neural network (CNN) to identify binding poses can also be applied to pose optimization. We also show that an iteratively-trained CNN that includes poses optimized by the first CNN in its training set performs even better at optimizing randomly initialized poses than either the first CNN scoring function or AutoDock Vina.
\end{abstract}

\section{Introduction}

A primary goal of structure-based drug design is to use the known three-dimensional structures of target proteins and ligands to discover new active compounds. One technique for this is virtual screening, in which a target structure is scored with molecules from a large database to identify potential actives \textit{in silico}. While orders of magnitude faster and cheaper than testing with experimental assays, virtual screening still has significant issues that limit hit rates (true actives per molecule screened) \cite{shoichet2004}. Virtual screening relies on docking to predict the conformation of each molecule when bound to the target in a stable complex, so its success is highly dependent on the accuracy of the binding poses and scores predicted by the underlying docking method.

A common approach to docking combines a scoring function with an optimization algorithm. The scoring function quantifies the favorability of the protein-ligand interactions in a single pose, which can be conceptualized as a point in a continuous conformation space. A stochastic global optimization algorithm is used to explore and sample this conformation space. Then, local optimization is employed on the sampled points, usually by iteratively adjusting the pose in search of a local extremum of the scoring function. Ideally, the scoring function is differentiable to support efficient gradient-based optimization. 

With the abundant structural data now publicly available in online resources like the Protein Data Bank \cite{Berman2000pdb}, machine learning finds a natural application in scoring function development for docking. In our previous work, we demonstrated that deep learning models, in particular convolutional neural networks, can effectively learn to discriminate between correct and incorrect binding poses when trained on three-dimensional protein-ligand structures \cite{ragoza2017cnn}. We also introduced an atomic grid representation for this task that ensured the convolutional neural network scoring function that was learned was differentiable with respect to atom positions.

In this paper, we extend our previous work with the following contributions:

\begin{itemize}

\item We train a convolutional neural network (CNN) scoring function to discriminate binding poses using the differentiable atomic grid format as input. Then, we optimize a set of randomized poses by maximizing the binding class output of the CNN using its gradient with respect to ligand atom positions.

\item We implement an iterative procedure where we extend the training set with the optimized poses and subsequently train an additional CNN scoring function with this extended training set. We also use the iteratively-trained CNN to optimize the randomized poses.

\item We compare poses optimized with a basic CNN and an iteratively-trained CNN as described above to poses optimized with AutoDock Vina. \cite{Trott2010}

\end{itemize}

\section{Related Work}

A fundamental question in structure-based drug design is how to quantify the complex chemical interactions between proteins and small molecules that correlate with binding \cite{shoichet2004}. Protein-ligand scoring functions have been developed specifically for this task. Scoring functions map from a three-dimensional conformation of a protein and ligand to a single value estimating the strength of the interaction. The various types of scoring functions are often motivated by empirical \cite{Bohm1994score, Eldridge1997, Wang2002xscore, Friesner2004glide, Korb2009plants, sminapaper} or knowledge-based \cite{Muegge1999, Gohlke2000, Mooij2005, Ballester2010, Huang2010, Zhou2011} strategies. These are generally simple functions of weighted terms designed to represent important noncovalent features of binding, like electrostatic interactions and hydrophobic contact. AutoDock Vina \cite{Trott2010} is a popular open-source docking program with a scoring function that combines both approaches--it has a physically-inspired functional form that was parameterized to fit experimental binding data.

Machine learning has emerged as an additional, increasingly popular approach to scoring function design. Algorithms including random forests, support vector machines, and neural networks have all been applied to scoring problems in various contexts \cite{Ashtawy2015, Sato2009, Ballester2010, zilian2013sfcscore, jorissen2005, schietgat2015predicting, sminapaper, deng2004,  chupakhin2013predicting, durrant2015ml, gonczarek2016, durrant2010nnscore,  durrant2011nnscore, wallach2015atomnet}. These have greater capacity to express nonlinear dependencies between chemical features \cite{li2014importance}. However, rigorous validation with carefully curated benchmarks is needed to mitigate sources of generalization error such as artificial enrichment and analogue bias \cite{Rohrer2009, Kramer2010, gabel2014beware, wallach2015atomnet}. In addition, most of these machine learning techniques still rely on feature engineering, which may omit factors of binding not fully understood or represented in the chosen features.

In computer vision, deep learning has superseded traditional feature-based techniques. Convolutional neural networks (CNNs) have revolutionized image recognition, as evidenced by the remarkable performance increases achieved by winners of the ImageNet Large Scale Visual Recognition Challenge in recent years \cite{imagenet, krizhevsky2012imagenet, simonyan2015vgg, szegedy2015going, msdeepresidual}. Near-human level image recognition performance has been reached with deep convolutional neural networks through innovations in model architectures and training practices \cite{simonyan2015vgg, szegedy2015going, msdeepresidual, srivastava2014dropout}. These models allow highly nonlinear functions to be learned directly from low-level data with minimal featurization. Instead, they infer a hierarchical structure of features that efficiently solves the training task \cite{lecun2015deep}.

Deep learning has also made significant progress in computational drug discovery. It has been applied both to the generation of molecular fingerprints and the multi-task prediction of cheminformatic properties based on them \cite{ramsundar:2015, DuvMacetal15nfp}. Convolutional neural networks trained on atomic grids have successfully learned scoring functions for binding prediction, pose discrimination, and affinity prediction \cite{ragoza2017cnn, wallach2015atomnet}. Other work has trained convolutional neural networks to predict binding affinities or quantum mechanical energies and forces without using an atom grid \cite{gomes2017, schutt2017}. These approaches instead use radially-pooled convolution filters on each atom. Thus, deep learning for drug discovery has mainly differed in how the chemical data are represented and in the type of scoring task.

Visualization has helped illuminate what deep neural networks learn to look for in their input, and in the process revealed some of their problematic tendencies \cite{mahendran2015understanding, simonyan2013deep, nguyen2015deep, goodfellow2015}. Activation maximization is an approach to visualization that uses gradient optimization on the input space \cite{mahendran2015understanding, simonyan2013deep}. When applied to an image recognition CNN, this generates novel images that emphasize what the model looks for in a particular class or feature it learned to identify. However, generated images do not appear natural unless constraints are imposed during the optimization process \cite{mahendran2015understanding}. Furthermore, this method allows the creation of adversarial examples--images that lie only slightly outside the training data distribution but nevertheless are incorrectly classified \cite{goodfellow2015}.

A currently unexplored application of deep learning to drug discovery is gradient-based optimization of chemical structures. Activation maximization performed on a neural network scoring function is analogous to local pose optimization, which is fundamental to molecular docking. A convolutional neural network with a differentiable input representation can therefore be used for both the scoring and optimization components of docking \cite{ragoza2017cnn}. Poses generated in this manner are susceptible to the same pitfalls of activation maximization in the image domain, so constraints are needed to ensure that the optimized poses are physically realistic.

\section{Methods}

In order to determine whether a CNN trained to discriminate binding poses could also effectively improve poses by gradient-based optimization, we utilized an iterative procedure involving two rounds of training and pose optimization. First, we trained a CNN to classify correct and incorrect binding poses and used it to optimize a set of randomly initialized poses. Then, we trained a second CNN that includes poses optimized by the first CNN in its training set. We used this iteratively-trained model to optimize the same initial random poses. Finally, we compared the poses optimized with each CNN to poses optimized with AutoDock Vina and assessed the ability of each method to improve randomized poses.

\subsection{Training set}

All datasets were derived from the 2016 version of the PDBbind general set \cite{Wang2005pdbbind}, which contains 13,308 complexes of ligands bound to target proteins from the Protein Data Bank \cite{Berman2000pdb} with structures determined by X-ray crystallography. Certain complexes were deemed unsuitable due to having unusually high energy or being unusually large. The internal energy of each ligand in the general set was calculated with Open Babel using the default MMFF94 force field. Those with energy greater than 10,000 kJ/mol were removed from the data set. The molecular weight was also calculated using Open Babel and ligands with weight greater than 1200 Da were removed. This resulted in a modified general set containing 12,482 structures. We then generated an initial training set by redocking the crystal structures with AutoDock Vina and taking approximately 20 of the top-ranked poses per target. We additionally included poses that resulted from local optimization of the crystal structure poses with AutoDock Vina, to ensure that each target had at least one correct binding pose in the training set while avoiding training directly on the crystal structures. Poses were labeled based on their root mean squared distance (RMSD) from the crystal pose. Any pose less than 2{\AA} RMSD from the crystal pose was labeled as a binding pose, and any pose greater than 4{\AA} RMSD was labeled as a non-binding pose. Poses between 2 and 4{\AA} RMSD were considered ambiguous, and were not included in the training set. The distribution of poses in the training set can be seen in Figure~\ref{init_rmsd_hist} and Table~\ref{training_set}.

\subsection{Random set}

\begin{figure}[tbp]
\centering
\includegraphics[width=0.7\linewidth]{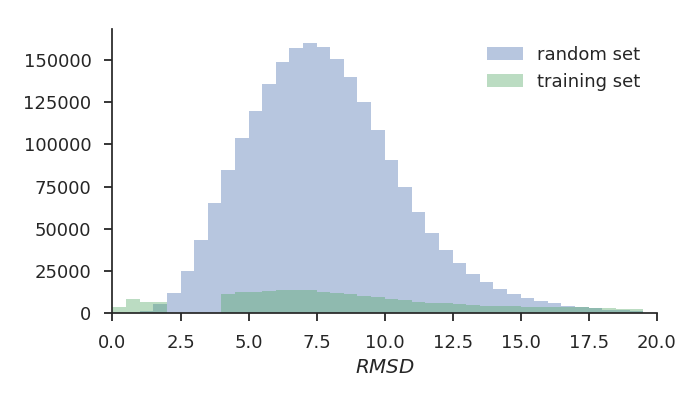}
\caption[]{\label{init_rmsd_hist} RMSD distributions of the poses in the training set, which were used to develop CNN scoring functions, and the poses in the random set that served as the initial poses for optimization.}
\end{figure}

In addition to the training set, we created a set of initial poses for optimization. This set consisted of the same 12,482 PDBbind structures as the training set, except that poses were generated by randomly sampling the conformation space of each ligand within a bounding box around the ligand in the crystal pose. We sampled 500 random poses per target, for a total of 6,241,000 poses. The RMSD distribution of these poses can be viewed in Figure~\ref{init_rmsd_hist}.

\begin{table}[tbp]
\centering
\begin{tabular}{ l | c | r }
               & per target & overall \\
  \hline \hline			
  binding poses     &  1.98 &  24,727 \\
  non-binding poses & 19.56 & 244,167 \\
  total poses       & 21.54 & 268,894 \\ 
\end{tabular}
\caption{\label{training_set} Distribution of binding and non-binding poses in the training set.}
\end{table}

\subsection{Model architecture and training}

We adopted the CNN architecture described in the model optimization results of \cite{ragoza2017cnn}, which consisted of a \texttt{MolGridData} layer, three modules of max pooling, convolution, and rectified linear activation layers, and finally a fully-connected layer with softmax output. The \texttt{MolGridData} layer used a 24\r{A} grid centered at the binding site with 0.5\r{A} resolution, and the same atom type scheme as in \cite{ragoza2017cnn} with the addition of a boron atom type. The convolution layers had 32, 64, and 128 filters respectively, and each had kernel size 3x3x3 and stride 1. The max pooling layers had kernel size 2x2x2 and stride 2. The fully-connected layer had 2 output classes, one for binding (low RMSD) and the other for non-binding (high RMSD), and the loss function was logistic loss with L2 regularization (\texttt{weight\_decay} = 0.001). The CNN was implemented in Caffe \cite{jia2014caffe} and trained by stochastic gradient descent (SGD) with momentum and an inverse learning rate policy (\texttt{base\_lr} = 0.01, \texttt{momentum} = 0.9, \texttt{gamma} = 0.001, \texttt{power} = 1). The training data order was shuffled each epoch and the classes were balanced within each batch of 50 poses. Additionally, each protein-ligand pose was rotated by a random amount and translated by a maximum of 2\r{A} along each axis prior to computing the atom grid. Models were trained for 100,000 iterations.

\subsection{Pose optimization}

In order to optimize poses using the gradient of the trained CNN, we need the partial derivatives of the CNN output (for the binding class) with respect to atomic coordinates. Formally, our CNN scoring function $f$ takes an atomic grid $\bm{G}$ (a vector of 3-dimensional grids of atom density called channels, one for each atom type) as input, and outputs a binding score. The atomic grid is produced from a list of atomic coordinates and atom types representing a pose. For each atom coordinate $\bm{a}$, its partial derivative of the output score is given by the chain rule:

\begin{equation}
\frac{\partial f}{\partial \bm{a}} = \frac{\partial f}{\partial \bm{G}} \frac{\partial \bm{G}}{\partial \bm{a}}
\end{equation}

Each atom only contributes density to the grid channel corresponding to its atom type, $\bm{G}_a$. The equation can be expanded into a summation over the grid points in that channel:

\begin{equation}
\frac{\partial f}{\partial \bm{a}} = \sum_{g \in \bm{G}_a} \frac{\partial f}{\partial g} \frac{\partial g}{\partial \bm{a}} \\
\end{equation}

The partial derivative of $f$ with respect to each grid point $g \in \bm{G}_a$ is calculated as part of a backward pass in Caffe, so we need the derivative of the value of a grid point with respect to an atom's position. The function that computes an atom's density at a grid point is a function of the distance $d$ to the grid point and the Van der Waals radius $r$ of the atom:

\begin{equation}
g(d, r) =
\begin{cases}
e^{-\frac{2{d}^2}{{r}^2}} & 0 \leq d < r \\
\frac{4}{e^2r^2}{d}^2 - \frac{12}{e^2r}d + \frac{9}{e^2} & r \leq d < 1.5r \\
0 & d \geq 1.5r \\
\end{cases} \\
\end{equation}

\begin{equation}
\frac{\partial g}{\partial d} =
\begin{cases}
-\frac{4d}{r^2}e^{\frac{-2{d}^2}{{r}^2}} & 0 \leq d \leq r \\
\frac{8}{e^2r^2}d - \frac{12}{e^2r} & r < d < 1.5r \\
0 & d \geq 1.5r \\
\end{cases} \\
\end{equation}

Therefore, one last application of the chain rule results in the final equation in terms of atomic coordinates by including the derivative of the distance function as an additional factor:

\begin{equation}
\frac{\partial f}{\partial \bm{a}} = \sum_{g \in \bm{G}_a} \frac{\partial f}{\partial g} \frac{\partial g}{\partial d} \frac{\partial d}{\partial \bm{a}} \\
\end{equation}

Applying this derivation, we computed the partial derivatives $\partial f/\partial \bm{a}$ for each pose in the random set and treated them as atomic forces affecting the ligand's translational, rotational, and internal degrees of freedom. Then the Broyden-Fletcher-Goldfarb-Shanno algorithm (BFGS) was applied to seek a local maximum of the binding class output in conformation space.  An early termination criteria was enforced whenever the improvement from one step of BFGS to the next was less than $1e{-5}$. We assessed the  RMSD from the crystal pose before and after optimization to establish whether it successfully improved the pose by bringing it closer to the pose observed in the crystal structure.

\subsection{Iterative training and pose optimization}

Since the training set poses were biased by what AutoDock Vina considered plausible, the CNN was not trained on examples of physically unrealistic poses. We expected that it might predict that such poses were favorable conformations and optimize towards them. To compensate for this, after the initial round of training and pose optimization, an extended training set was constructed by augmenting the original training set with the poses generated by optimizing the random set with the CNN. This extended training set contained both Vina-docked poses and CNN-optimized random poses. A second CNN was trained on this extended training set, using the exact same training procedure as the first. Then the random set poses were optimized with the second CNN, again using the exact same optimization procedure as the first round. In the results and discussion, the first CNN is referred to as CNN1 and the iteratively-trained one is referred to as CNN2.

\section{Results}

We evaluated the results of pose optimization using CNN1 and CNN2 and compared them with AutoDock Vina. None of the three methods consistently improved the random set poses by a large amount. Both Vina and CNN1 had positive mean change in RMSD ($\Delta$RMSD), while CNN2 had a small negative mean $\Delta$RMSD.  The distributions of $\Delta$RMSD for each method had distinct shapes, and there were differences in the joint distributions of $\Delta$RMSD with initial RMSD.

\begin{figure}[tbp]
\centering
\includegraphics[width=0.7\linewidth]{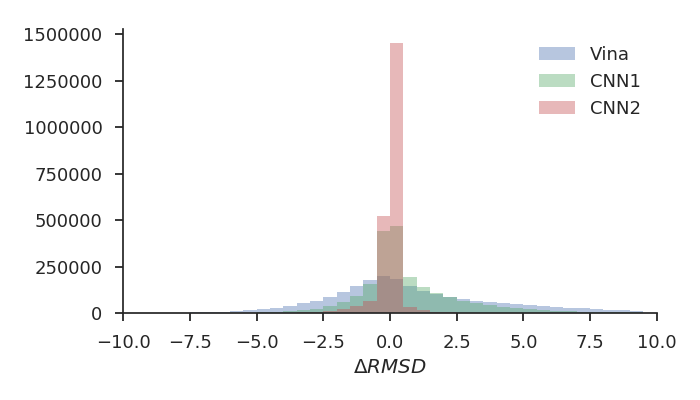}
\caption[]{\label{delta_rmsd_hist} Distributions of $\Delta$RMSD from crystal pose after optimizing the random set poses using Vina, CNN1, and CNN2.}
\end{figure}

\begin{table}[tbp]
\centering
\begin{tabular}{ l | c | c | c }
                    & Vina & CNN1 & CNN2 \\
  \hline \hline	
  all poses         & 0.902 $\pm$ 0.002 & 0.774 $\pm$ 0.002 & \textbf{-0.071 $\pm$ 0.000} \\ 
  binding poses     & 0.595 $\pm$ 0.030 & 2.031 $\pm$ 0.038 & \textbf{0.103 $\pm$ 0.012} \\
  ambiguous poses   & 0.925 $\pm$ 0.007 & 1.497 $\pm$ 0.007 & \textbf{-0.254 $\pm$ 0.002} \\
  non-binding poses & 0.901 $\pm$ 0.002 & 0.719 $\pm$ 0.002 & \textbf{-0.059 $\pm$ 0.000} \\

\end{tabular}
\caption{\label{opt_results} Mean $\Delta$RMSD from crystal pose for each of the optimization methods shown with standard error of the mean. Results are shown for all poses of the random set, as well as for subsets of poses based on initial RMSD. Binding poses were initially less than 2{\AA} RMSD, ambiguous poses were between 2 and 4{\AA} RMSD, and non-binding poses were greater than 4{\AA} RMSD.}
\end{table}

Optimization of poses with Vina and CNN1 did not tend to decrease the RMSD from the true binding poses. Table~\ref{opt_results} shows that both methods increased the RMSD of randomly initialized poses on average, though CNN1 did so to a lesser extent than Vina ($\mu$ = 0.774 for CNN1 and $\mu$ = 0.902 for Vina). CNN2 was the only optimization method that decreased the RMSD of random poses on average ($\mu$ = -0.071). As seen in Figure~\ref{delta_rmsd_hist}, the $\Delta$RMSD for all three methods of pose optimization is centered near zero, but the shapes of the distributions vary. The Vina distribution is the most dispersed ($\sigma$ = 3.445) and also the most symmetrical. The distribution of poses optimized with CNN1 has a higher peak around zero, less variance ($\sigma$ = 2.346) and more right skewness. The distribution of $\Delta$RMSD for CNN2 is the most concentrated of any method ($\sigma$ = 0.515), with nearly all poses only slightly increasing or decreasing in RMSD. The CNN2 distribution displays some left skewness as well.

The differences in pose optimization by each method are clarified further by considering the impact of initial RMSD. The poses can be categorized based on the labels used for constructing the training set: binding poses were initially less than 2{\AA} RMSD, non-binding poses were greater than 4{\AA} RMSD, and all other poses were ambiguous. As displayed in Table~\ref{opt_results}, Vina and CNN1 both increased the RMSD on average for all categories of poses, but exhibit opposite trends. Vina had the lowest mean $\Delta$RMSD for binding poses ($\mu$ = 0.595) and was worse at optimizing ambiguous and non-binding poses ($\mu$ = 0.925 for ambiguous and $\mu$ = 0.901 for non-binding). On the other hand, CNN1 performed the worst on binding poses ($\mu$ = 2.031) and comparatively better on ambiguous ($\mu$ = 1.497) and non-binding poses ($\mu$ = 0.719). Despite that CNN1 performed better than Vina overall, the advantage is only for non-binding poses--it did significantly worse than Vina when optimizing binding and ambiguous poses. CNN2 outperformed CNN1 and Vina in every category, and binding poses are the only ones that it did not improve on average ($\mu$ = 0.103). It was best at decreasing the RMSD of ambiguous poses ($\mu$ = -0.254) but it slightly decreased RMSD for non-binding poses as well ($\mu$ = -0.059). It is noteworthy that the overall best-optimized poses were ambiguous ones optimized with CNN2, especially since ambiguous poses were the hardest to optimize for Vina.

\begin{figure}[tbp]
\centering
\includegraphics[width=\linewidth]{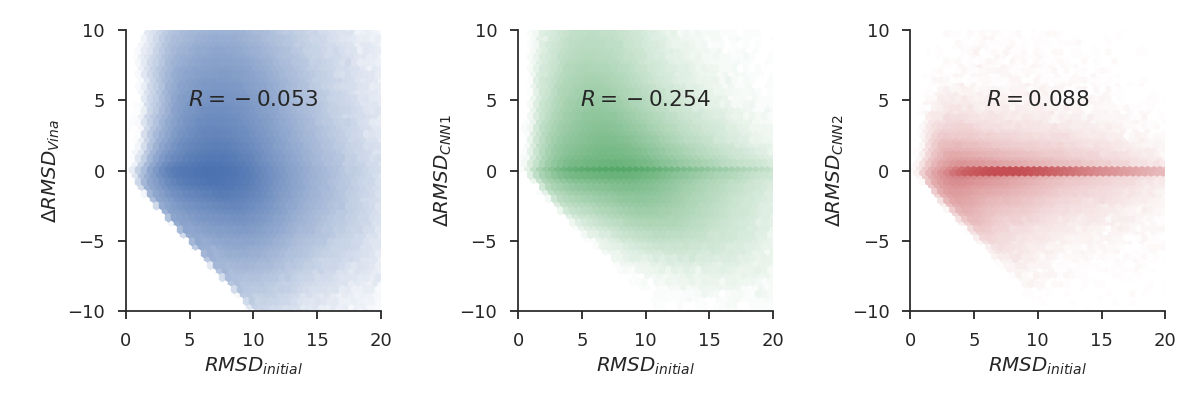}
\caption[]{\label{delta_rmsd_corr} Correlation of $\Delta$RMSD with initial RMSD for random set poses optimized by Vina, CNN1 and CNN2. }
\end{figure}

The effect of initial RMSD on $\Delta$RMSD is further analyzed in Figure~\ref{delta_rmsd_corr}. None of the methods show strong correlations between the $\Delta$RMSD and the initial RMSD, but the joint distributions have interesting characteristics. Both CNN1 and CNN2 display the expected strong peak at zero $\Delta$RMSD that is absent for Vina, but the peak is not constant with respect to initial RMSD. It appears to fade as the initial RMSD approaches zero for both of these methods, which would correspond with higher variance. This entails higher likelihood of CNN optimization significantly altering a pose if it starts closer to the crystal structure.

\section{Discussion}

\begin{figure}[tbp]
\centering
\includegraphics[width=.33\textwidth]{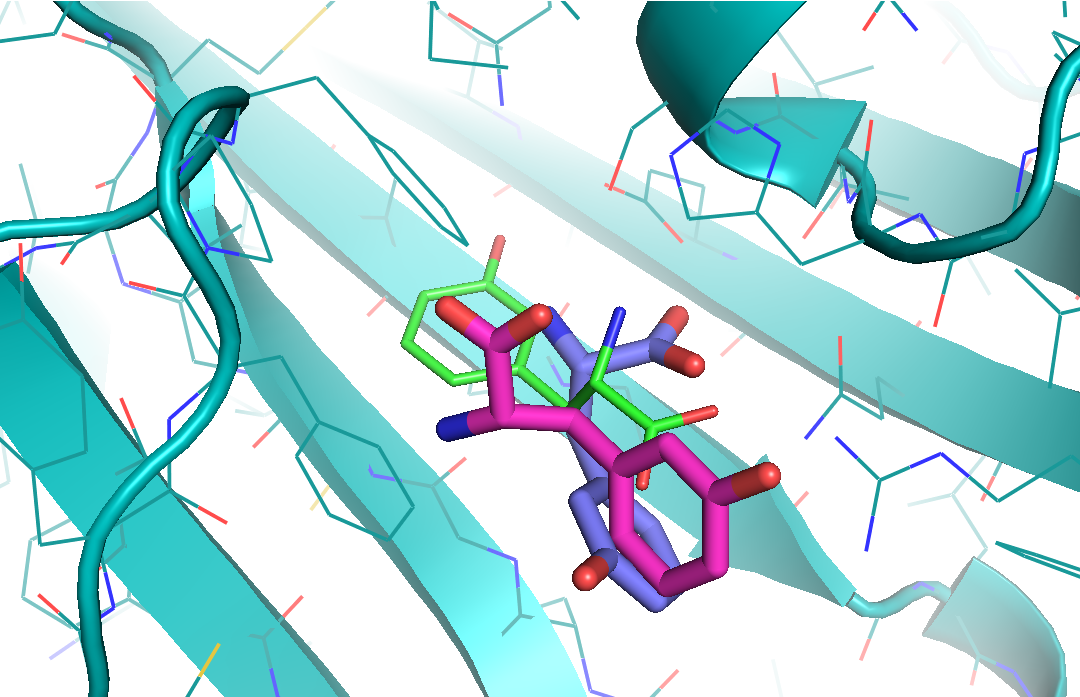}\hfill
\includegraphics[width=.33\textwidth]{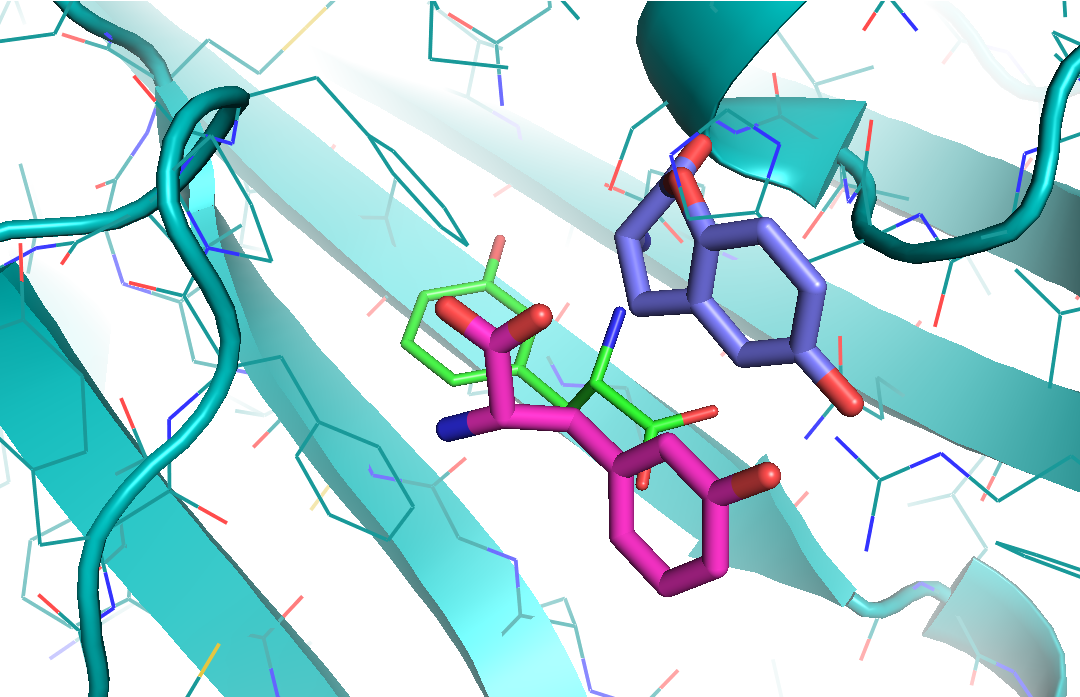}\hfill
\includegraphics[width=.33\textwidth]{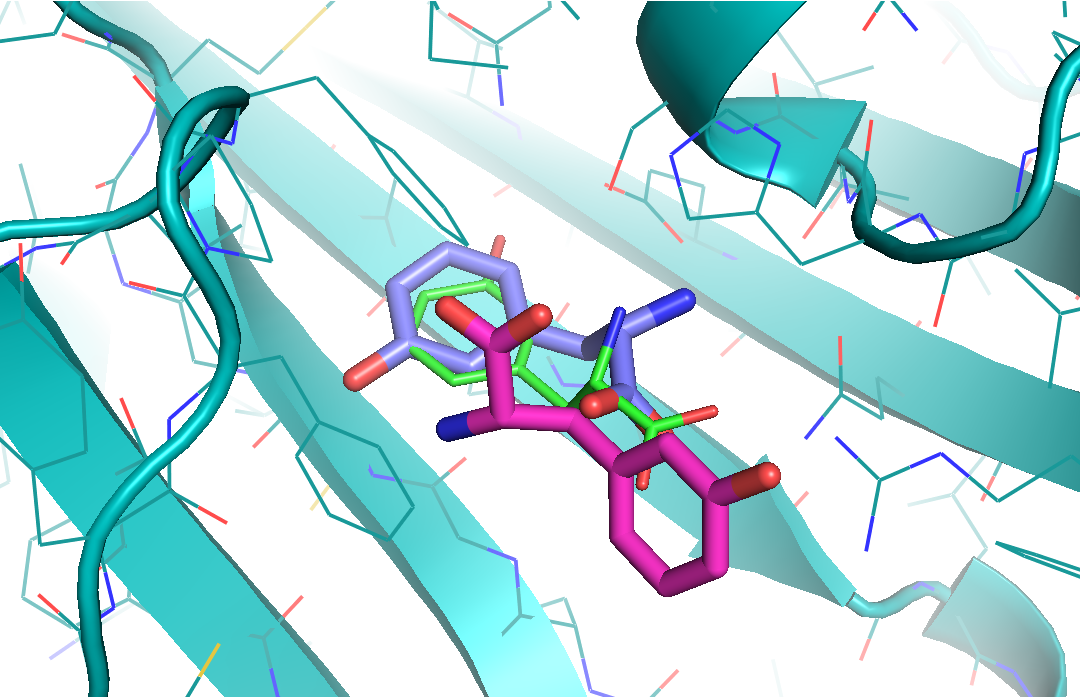}
\caption{A randomized pose of target 3HFV optimized with Vina (left), CNN1 (middle), and CNN2 (right). The crystal pose is shown in green, the initial pose is shown in magenta, and the optimized poses are blue.}
\label{3hfv_7}
\end{figure}

While none of the methods effectively optimized the majority of the random set poses, there is reason to believe that the iterative training procedure solved some of the issues that prevented optimization in CNN1. Pose optimization with a CNN depends on the decision boundary the model learns during training, which in turn depends on the distribution of the training set. The original training set was comprised only of crystal poses redocked with Vina. These poses were biased towards certain physically realizable regions of the conformation space, owing to the physically-inspired form of the Vina scoring function. Therefore, CNN1 was not trained to recognize poses that would be considered physically improbable by Vina, leaving important regions of the conformation space undetermined. This lead to some of the poses moving in physically unrealistic directions when optimized by CNN1, only limited by the fact that we optimized on the ligand degrees of freedom rather than allowing atoms to move completely independently along their gradients. An example of this is seen in Figure~\ref{3hfv_7}, where the pose optimized by CNN1 overlaps itself as well as the protein. We anticipated these non-physical optimized poses--one of the goals of iterative training was to see if CNN2 would learn to account for important features like steric effects that were lacking in the Vina-docked poses. In Figure~\ref{3hfv_7}, CNN2 successfully avoids the steric clashes that CNN1 produced, and instead manages to flip the pose to its correct orientation and bring it close to the correct binding conformation.

Iterative training may have helped the CNN to better define the regions of conformation space not covered by the original training set, but this also prevented the model from significantly altering the poses. This is evidenced by the reduction in variance of the distribution of $\Delta$RMSD from CNN1 to CNN2. The score surface of CNN1 was probably rougher in the regions of conformation space where the random poses  were found. This would imply that the random poses had larger gradients for CNN1, which lead to larger magnitude changes in RMSD. Since CNN2 was trained on these poses, it predicted them with higher confidence, leading to a smoother, flatter score surface in the vicinity of the non-ambiguous random poses. This explanation is consistent with the theoretical characteristics of the training task. The CNNs were trained using a softmax logistic loss in which the only region of conformation space that does not induce an extrema in the softmax logistic loss layer is between 2 and 4{\AA} RMSD. The more generally accurate the score predictions of such a model, the sharper the decision boundary, and the closer the gradients are to zero for any pose not close to the decision boundary. This hypothesis is supported by data in Table~\ref{opt_results} showing that the poses that were optimized best by CNN2 were the ambiguous ones. The peak around zero $\Delta$RMSD by CNN1 and CNN2 for high initial RMSD poses observed in Figure~\ref{delta_rmsd_corr} can be explained by this theory as well, since these poses would be far away from the decision boundary and therefore have small gradients. It seems that the scoring and optimization tasks as we defined them are mutually incompatible for the broadest class of poses.

In order to improve pose optimization with a CNN scoring function, a different training task for classifying poses would be beneficial. For example, a network could be trained to predict the RMSD directly, a different loss function could be used, or the binary classification could otherwise be modified to incorporate some notion of the energy landscape. The resulting scoring function would ideally be less prone to vanishing gradients, and potentially better at optimizing poses greater than 4{\AA} RMSD from the crystal structure. Another important point to remember regarding our method is the insight from \cite{nguyen2015deep} that the iterative training approach is not enough to eliminate adversarial examples. Though CNN2 may have successfully smoothed the regions of conformation space that were indeterminate for CNN1, it is feasible that CNN2 would not generalize to a different set of randomized poses, or a set of poses generated by some other sampling method. Rather than continuously extending the training set with optimized examples, the solution put forth in computer vision is to incorporate the generation of novel training examples into the architecture of the model itself. This is accomplished by simultaneously learning a discriminative model and a generative model in what is known as a generative adversarial network (GAN) \cite{goodfellow2014gan}. A future direction for this work is to apply a GAN to pose prediction. Such a model could potentially learn not only to discriminate binding poses, but to generate realistic atom grids from scratch. This could be a new way of sampling poses, but it could also go beyond that to the \textit{de novo} design of ligands that optimally activate the scoring function.

This work represents a first attempt at generating optimal ligand binding poses using the gradient of a convolutional neural network scoring function. Though the results imply that changes to the training procedure are needed before these models can be effectively incorporated into off-the-shelf docking software, there is enormous potential for further exploration and improvement. We plan to continue the investigation of deep learning for computational drug discovery, and in the hopes that others do as well we include all of our code and models as part of gnina, our open-source docking software, available at \url{https://github.com/gnina}.

\section*{Acknowledgements}

This work is supported by R01GM108340 from the National Institute of General Medical Sciences and by a GPU donation from the NVIDIA corporation.

{\footnotesize
\bibliography{biblio}
}

\end{document}